\documentclass[conference]{IEEEtran}
\usepackage{blindtext, graphicx}
\usepackage[utf8]{inputenc}
\usepackage[spanish, activeacute]{babel} %Definir idioma español
\usepackage{listings}
\usepackage{amssymb}
\usepackage{soul}
\usepackage[breaklinks=true]{hyperref}

\begin{document}
\title{A Novel Perception and Semantic Mapping Method \\for Robot Autonomy in Orchards}

\author{
Yaoqiang Pan$^{1}$, Hao Cao$^{1}$, Kewei Hu$^{1}$, Hanwen Kang$^{1,*}$, Xing Wang$^{2,*}$\\ \\
\small{$^{1}$ College of Engineering, South China Agriculture University, China}\\
\small{$^{2}$ Robotics and Autonomous System, lData61, CSIRO, Australia}\\
\small{$^{*}$ Correspondence Authors}\\
}
\maketitle

\begin{abstract}
Agricultural robots must navigate challenging dynamic and semi-structured environments. Recently, environmental modeling using LiDAR-based SLAM has shown promise in providing highly accurate geometry. However, how this chaotic environmental information can be used to achieve effective robot automation in the agricultural sector remains unexplored. In this study, we propose a novel semantic mapping and navigation framework for achieving robotic autonomy in orchards. It consists of two main components: a semantic processing module and a navigation module. First, we present a novel 3D detection network architecture, 3D-ODN, which can accurately process object instance information from point clouds. Second, we develop a framework to construct the visibility map by incorporating semantic information and terrain analysis. By combining these two critical components, our framework is evaluated in a number of key horticultural production scenarios, including a robotic system for in-situ phenotyping and daily monitoring, and a selective harvesting system in apple orchards. The experimental results show that our method can ensure high accuracy in understanding the environment and enable reliable robot autonomy in agricultural environments.
\end{abstract}

\section{Introduction}
Robot autonomy has become a key aspect of precision agriculture, which leverages the emerging robotics, sensor, and AI technologies to automate the process of scientific data collection \cite{kang2020fast, fu2020application}, yields estimation\cite{liu2023orb, zhou2020real}, fruit growth monitoring\cite{oliveira2021advances}, and fruits harvesting\cite{wang2023development, kang2022accurate}. 
Amidst the varied subsystems of an intelligent agricultural robot, encompassing machine vision \cite{patel2012machine}, end-effector \cite{baohua}, mobile manipulation \cite{au2023monash}, and others, navigation emerges as a pivotal element, empowering the robot for autonomous operations in unstructured agricultural environments, such as orchards and plantation \cite{zhou2022intelligent, gul2019comprehensive}. 
In summary, an accurate and effective representation of environments is key to achieving robot autonomy.
\begin{figure}[h]
    \centering
    \includegraphics[width=.45\textwidth]{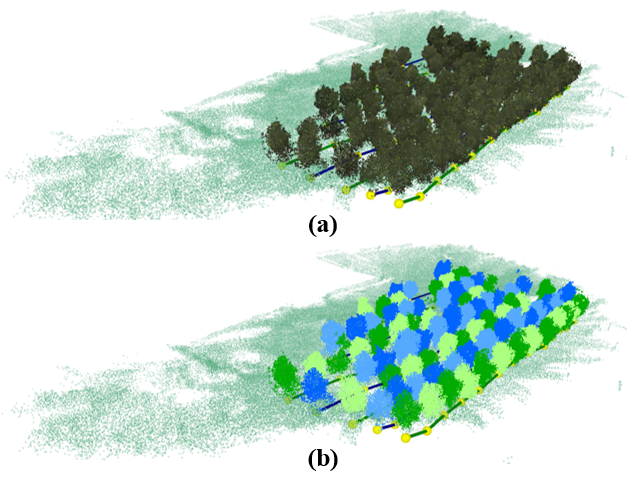}
    \caption{Mapping and semantic mapping results of orchards, (a) colourised map, (b) semantic map of orchards.}
    \label{fig: semantic mapping}
\end{figure}

Simultaneous Location And Mapping (SLAM), an emerging technique widely used in robot self-positioning and mapping, shows its huge potential in many agricultural automation applications. It uses sensors such as cameras, Light Detection And Range (LiDAR), and radar to acquire external information from the environment. At the front end of the SLAM, this information is used to compute the odometry, which is used to construct the map of the environment. At the back end of SLAM, the map of the environment is fine-tuned through close-loop detection and overall map optimisation. Although SLAM provides robots with a strong ability to create the right geometry map for implementing robot automation in many scenarios, it lacks the ability to process the information within the map, which limits the reliability and effectiveness of the robot in many conditions, especially in the sometimes challenging horticultural environments. Having a semantic understanding of the environment can significantly increase the capability of the robotic application. For example, in an orchard, if the robot knows the layout of a semi-structured orchard and the location of each tree, it can automatically find its way to the given position and finish the job, rather than having a human click on the screen each time to tell the robot exactly where to go.

The increasing demand for robotic automation in agriculture places high demands on the robot's ability to understand its environment. To achieve this goal, robots need to extract information from the scene and understand their surroundings on the map. That is, beyond the pure geometry map constructed from SLAM, a semantic map must be created to represent the semantically meaningful instance-level information and traversable regions. Deep learning, a powerful tool for processing and extracting semantic information from sensor inputs, can play a key role in this process. However, although significant progress has been made in the construction of semantic maps using 2D image data \cite{wang2022geometry}, semantic processing models directly on 3D point cloud data have not been widely explored. Compared with semantic processing on 2D image data, semantic processing on 3D point cloud data can directly use the output of SLAM and does not require any additional calibration or data format conversion, which avoids numerical errors and large computational consumption. However, due to the unstructured and sparse nature of 3D point clouds, efficient processing of semantic information within point clouds remains a challenge.  

This work focuses on addressing the two aforementioned challenges: (1) accurate semantic processing of the geometric map derived from SLAM, and (2) effective construction of a semantic map for robotic automation. In this work, by taking advantage of LiDAR-based mapping methods, a novel framework including a semantic processing method and a semantic-aware navigation method is proposed to achieve effective and reliable robotic autonomy in orchards. First, a novel 3D detection network is presented to extract instance-level semantic information from point cloud maps. Second, taking advantage of the semantic processing capability, we propose a navigation module that enables effective and reliable global path planning by constructing a terrain-aware visibility graph map. Specifically, our contributions are as follows.
\begin{itemize}
    \item Create a fast and precise 3D Object Detection Network, 3D-ODN, which can process 3D point cloud and extract instance-level information from map.
    \item Develop a novel semantic mapping framework for semi-unstructured orchards modeling by using semantic processing and terrain analysis.
    \item Demonstrate the proposed semantic mapping and navigation framework on mobile robots in multiple orchards, covering a viriety of horticulture scenarios.
\end{itemize}
The rest of this paper is organised as follows. Section \ref{section: review} surveys related work, followed by the proposed methodologies in Section \ref{section: method}. Section \ref{section: overview} overviews the system setup of our approach, Section \ref{section: network} introduces the architecture and implementation of the 3D-ODN, while Section \ref{section: Accessible corridors} introduces the method of the semantic mapping framework. The experimental results and discussions are given in Sections \ref{section: expriment}. and then the conclusions are given in Section \ref{section: conclusion}. 
\begin{figure*}[ht]
    \centering
    \includegraphics[width=.995\textwidth]{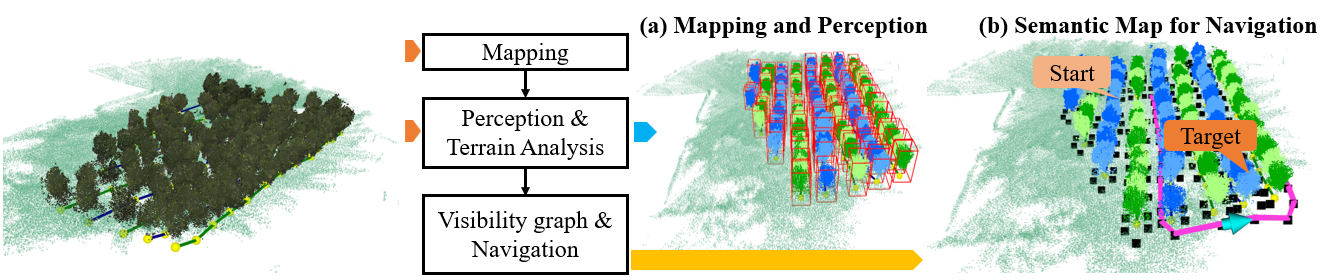}
    \caption{System-overview (a) Mapping and Perception (b) Semantic Map for Navigation.}
    \label{fig:system}
\end{figure*}

\section{Related Works} \label{section: review}
\subsection{Review on Semantic Mapping}
Semantic information plays a crucial role in advancing concurrent robotic SLAM systems, as demonstrated by Bowman et al. \cite{rw1}, who introduced a probabilistic data association-based approach that integrates inertial, geometric and semantic observations for the continuous pose, discrete data association and semantic label optimisation. Ouyang et al. \cite{rw7} proposed an interactive SLAM method that optimises point cloud poses through segmentation, while Bavle et al. \cite{rw4} presented a real-time visual semantic SLAM framework that uses low-level odometry and planar geometric information for sparse semantic mapping. In addition, Li et al. \cite{rw5} presented SA-LOAM, a closed-loop semantic LiDAR SLAM based on LOAM that uses semantics for matching, downsampling and planar constraints to construct globally consistent semantic maps, highlighting the integral role of semantic information in these semantic SLAM approaches. While previous approaches to SLAM have incorporated semantic information for localisation and mapping, they have fallen short of constructing dense maps conducive to autonomous navigation, Semantic Mapping enriches SLAM systems by integrating object categories, properties, and states through semantic segmentation and recognition, allowing robots to generate more holistic and meaningful maps that take into account both geometric structure and semantic information for improved environmental understanding and interaction. 

Zaenker et al. \cite{rw12} developed an algorithm to generate semantic layers from RGB-D images, creating a semantic map detailing the locations and labels of objects in the environment. Their method incrementally estimates object distribution by integrating semantic information from new image frames. However, its reliance on distance calculations between targets for object identification poses challenges in scenarios with high object similarity, such as orchards. In the field of off-road autonomous driving for all-terrain vehicles (ATVs), Maturana et al. \cite{rw13} proposed a semantic mapping system that aims to construct a 2.5D grid map with terrain elevation and semantic information. While providing semantic maps, this approach represents semantic information as a collection of points, lacking the high-dimensional topological detail required for sophisticated planning. Narita et al. \cite{rw11} presented Panoptic-Fusion, an online semantic mapping system based on object-to-object mapping. The system extracts semantic information from image frames using Mask R-CNN and maps it onto the camera depth point cloud. Despite its ability to construct a semantic map, its dependence on camera RGBD information limits its application to large outdoor scenes. Despite these advances, these methods are not explicitly designed for agricultural and horticultural contexts, which hinders the adaptability and reliability of robotic automation in these environments.

\subsection{Review on 3D Semantic Processing Methods}
While significant progress has been made in object detection within two-dimensional images by extracting semantic information, the field of three-dimensional point cloud-based object detection remains a challenging and open area of research \cite{lin2020color}. Unlike 2D images, which may lack detailed spatial distribution information, 3D point clouds obtained by SLAM methods provide a more comprehensive representation of object distribution, especially in complex scenes with occlusions and overlaps. This growing recognition of the potential of 3D point clouds has led to novel approaches to object detection. Yu et al. \cite{rw8} presented an approach that uses an improved F-PointNet and a three-dimensional clustering method for the precise positioning of ripe pomegranate fruits. This method combines RGB-D feature fusion mask R-CNN for fruit detection and segmentation, integrates PointNet with the OPTICS algorithm for point cloud segmentation, and uses sphere fitting to determine the size and location of the pomegranates. Chen et al. \cite{rw9} proposed a deep learning framework for direct processing of forest point clouds to achieve Individual Tree Crown (ITC) segmentation. The method involves voxel subdivision, PointNet for voxel-scale crown identification, and the use of highly correlated gradient information to delineate canopy boundaries. In addition, Wei et al. \cite{rw10} presented BushNet, a novel point cloud segmentation network that incorporates a minimum probability random sampling module for efficient sampling of large point clouds and a local multidimensional feature fusion module to enhance sensitivity to shrub point cloud features, thereby addressing the inherent challenges of 3D object detection. Our work introduces the 3D-ODN model, which innovatively combines space subdivision, feature coding, and a detection network to efficiently estimate the bounding boxes of orchard elements, in particular fruit trees, by extracting geometric features in a Bird's Eye View (BEV) and transforming 3D point cloud maps into 2D pseudo-images for accelerated and effective object detection.

\section{Methods} \label{section: method}

\subsection{System overview} \label{section: overview}
The framework of the proposed perception and semantic mapping framework is shown in Figure \ref{fig:system}. The framework can be divided into three steps, which are mapping, perception and semantic processing on a map. First, the orchard data is collected by a mobile robot with LiDAR, either in a tel-controlled or autonomous manner. The mapping process is carried out using LIO-SAM mapping algorithms on the collected data. Afterwards, the fruit trees of the orchards can be detected and localised by the 3D-ODN model using the geometric information within the map. Next, a semantic processing module was developed. This module extracts the topography by Cloth Simulation Filter (CSF) and combines the prediction results of 3D-ODN to extract the topological information of the point cloud map and construct the semantic map. Finally, the constructed semantic map includes information about each fruit tree that can be used to plan autonomous operations and a node connection map that can navigate robots to access each fruit tree within the map.

\subsection{Mapping Method}
The mapping process is performed using the LIO-SAM \cite{shan2020lio}. In mapping, RS Helios and external IMU (400 Hz) are used as mapping data sources for LIO-SAM. We do not use GPS in LIO-SAM surveying because obstacles in the orchard can affect the stability and accuracy of GPS signals. LIO-SAM uses the measurements from the IMU to infer the motion of the robot. The estimated motion from the IMU pre-integration corrects the point clouds and provides an initial estimate for the LiDAR odometry optimisation. The LiDAR odometry solution is then used to estimate the IMU bias. LIO-SAM can effectively simulate and describe the geometric characteristics of orchards. For presentation purposes, this orchard is coloured in this paper using camera data. The colour map of the orchard is shown in Figure \ref{fig: semantic mapping} (a) and Figure \ref{fig: semantic mapping} (b) is the semantic map of the orchard extracted by the method in this paper. We can see that the global map of the orchard is very unstructured. Meanwhile, the points within the map are very unevenly distributed.  

\subsection{Perception Model} \label{section: network}
\begin{figure}[ht]
    \centering
    \includegraphics[width=.48\textwidth]{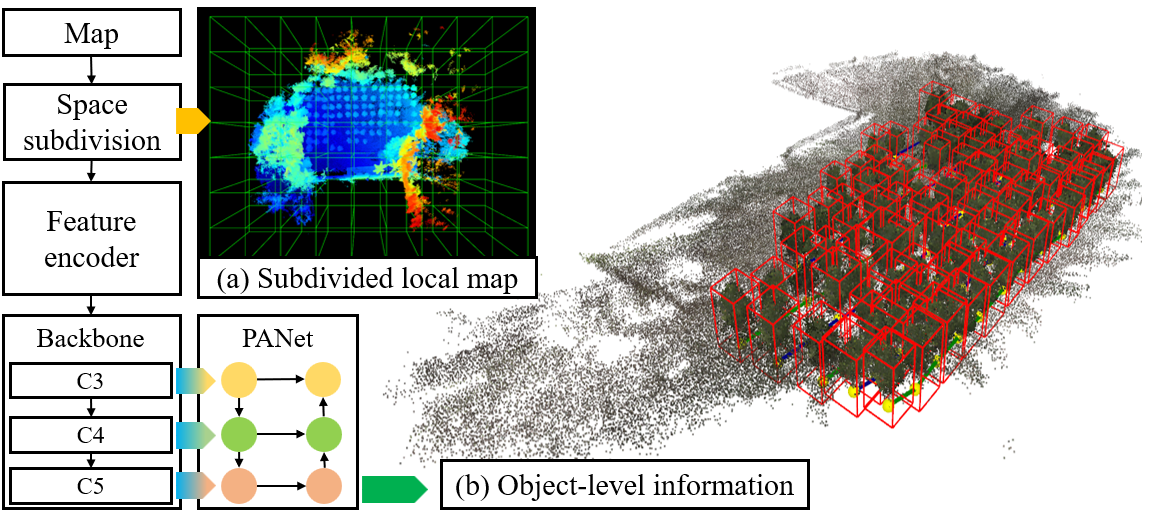}
    \caption{Network architecture of tree detection.}
    \label{fig: network}
\end{figure}
\textbf{3D-ODN}: This section presents a novel 3D perception model designed for a single-stage detection network operating on point cloud maps. The proposed network structure is shown in the Figure \ref{fig: network}. The proposed 3D-ODN model consists of three key steps: space subdivision, feature coding, and object detection. First, the entire map is segmented into equidistant and uniformly sized local maps. Each segmented local map is then processed through a feature encoder to extract geometric features from a BEV perspective and transform 3D point cloud data into 2D feature maps. These 2D feature maps are then fed into a detection model, which uses a modified Yolov7 architecture with the backbone network adapted to accommodate the feature map output from the encoder. The role of the backbone network is to extract features, and the Path Aggregation Network (PANet) \cite{liu2018path} modules combine feature maps from different layers for improved prediction. In particular, our innovation lies in the efficient conversion of 3D point cloud images into 2D feature maps, which improves the accuracy of object detection. The final prediction results are then remapped from 2D back to the original point cloud map, resulting in accurate prediction boxes for fruit tree objects. This approach streamlines the perception process and utilises a single-stage detection network for robust 3D object detection.

\textbf{Space Subdivision}: To address the challenges of efficiently processing geometric details within 3D geometry using a space subdivision strategy, a 10m local map is extracted from the global map. However, this approach poses a challenge as some fruit trees may only partially appear in the local map due to the space subdivision. To solve this problem, we introduce a sliding window detection method. After detecting a 10$m$ x 10$m$ local point cloud map, the window is incrementally shifted by 8$m$ in the positive x-axis direction at each step to ensure full coverage of fruit trees not included in the previous local map. The maximum forward sliding distance is set to 2$m$, allowing effective coverage by adjusting the window to 8$m$ (10$m$ minus the canopy diameter). This sliding window procedure is iterated along the x and y axes until the entire global map has been traversed. In cases where fruit trees span several subdivided local maps, NMS is used to retain the predictions with the highest confidence and eliminate overlapping predictions.

\textbf{Feature Encoding}: To leverage a 2D detection network to process point cloud information, we first convert a point cloud to a 2D pseudo image. That is, the local point cloud will be transferred to the BEV angle and divided into evenly-spaced grids on the x-y plane. This step creates a list of pillars with the ultimate extent on the z-axis direction. The points within each pillar will be used to extract features of the local geometry. The feature indicators include density, average height, and the angle between the principal vector and the z-axis. The dimensions of each pillar in the x and y directions are the side length/corresponding number of grids in the x and y directions of the local map. For example, if a 10$m$ × 10$m$ local map is feature extracted at a resolution of 128 × 128, the size of each grid is (10$m$/128) × (10$m$/128), and the area occupied is 0.00610 $m^2$. Some pillars are empty or only have a very small number of points due to the sparsity of the point cloud. Therefore, we apply a requirement that only the pillars with a point number large than 100 will be used to extract features. In this way, a 3D point cloud is converted into a 2D pseudo image. 

\textbf{Network Architecture}: We leverage a one-stage detection network architecture to perform object-level information perception on local maps. The network utilises ResNet-50 as the backbone. The input to the backbone is a 2D pseudo-image obtained by transforming the local point cloud map through the feature encoder. Then, the feature maps from the C3, C4, and C5 levels of the backbone are used to construct the multi-scale feature pyramid network. In this work, the PANet is used here to enhance the multi-scale image feature extraction and processing. Then, the processed feature maps from the PANet are used to perform detection. The combined detection results from the C3, C4, and C5 level of the PANet forms the predicted objects. Finally, NMS is employed to filter the predictions and generate the ultimate prediction results in the form of a 2D pseudo-image. For predictions made on the 2D pseudo-image, they are projected onto the x and y directions of the point cloud map, while the highest and lowest points within the x and y range of the point cloud are selected to define the boundaries in the z-direction for the fruit trees. This process ultimately yields the predictions on the three-dimensional point cloud map. 

\subsection{Semantic Mapping and Navigation} \label{section: Accessible corridors}
\begin{figure}[h]
    \centering
    \includegraphics[width=.49\textwidth]{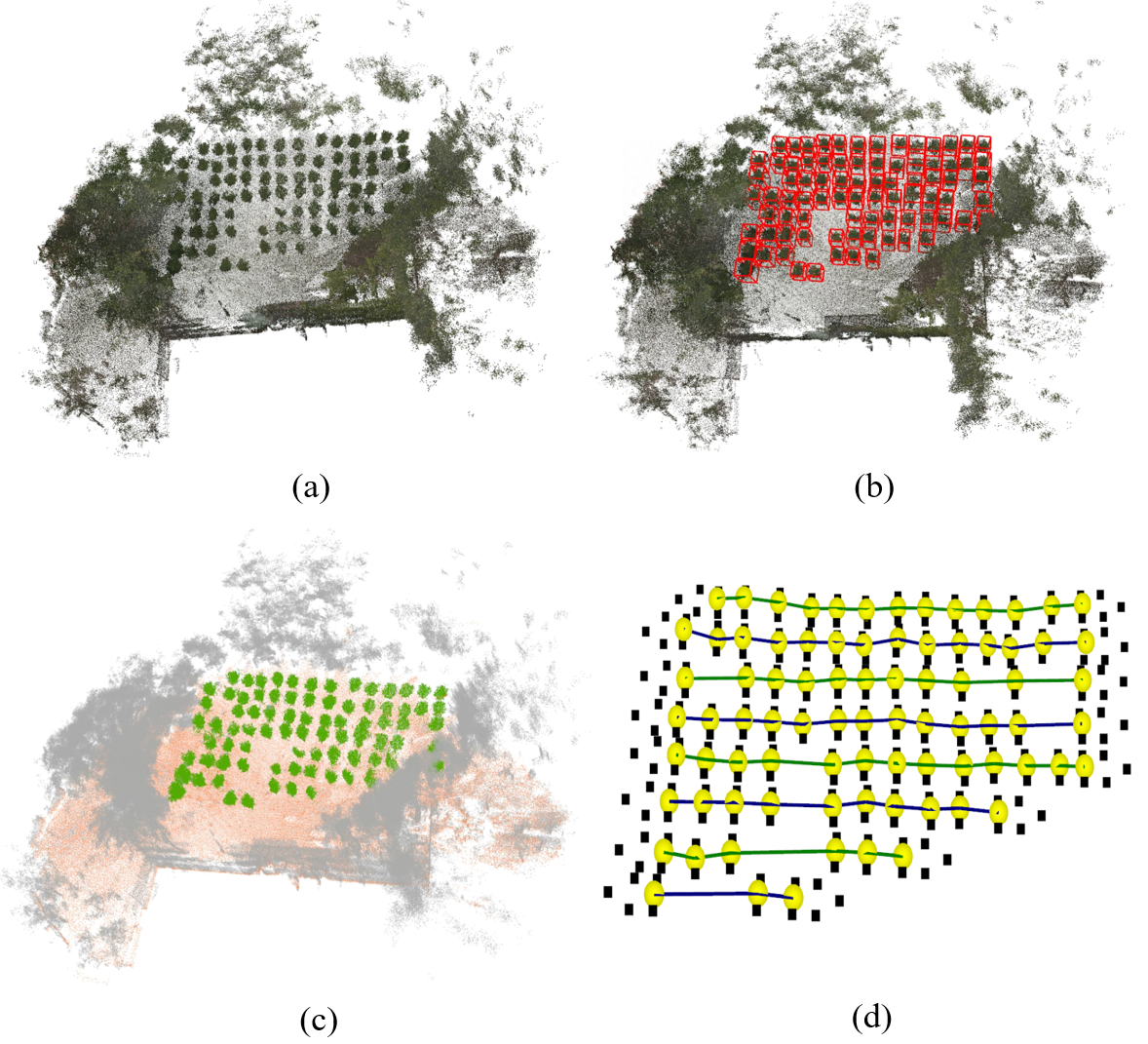}
    \caption{Semantic mapping processing (a) original point cloud map (b) perception results (c) semantic mapping results (d) visibility graph map.
}
    \label{fig: Traversability analysis process}
\end{figure}

\textbf{Traversability analysis}: The acquired map is then processed by the perception and terrain analysis. That is, we first utilise the 3D-ODN model to perform detection on the global map, as shown in Figure \ref{fig: Traversability analysis process} (b). This step will generate the list of detected fruit trees. Then, a terrain analysis method based on the CSF algorithm \cite{CSF} is utilised here to find the ground and obstacles of the orchards. The CSF algorithm filters ground points by simulating a virtual cloth dropped onto an inverted point cloud. Modeled as a grid of interconnected particles with mass, this virtual cloth effectively separates ground points. The relationship between particle positions and forces is governed by Newton's second law, expressed by Eqs. \ref{CSF}.
\begin{equation}\label{CSF}
    m\frac{\partial X(t)}{\partial t^2} = F_{gravity}+F_{spring}+F_{damping}
\end{equation}

In this way, the points of the global map are separated into traversable terrains and obstacles. By further combining the perception results of fruit trees, the points of the global map are classified into three classes: fruit trees, traversable terrains, and other obstacles, as shown in Figure \ref{fig: Traversability analysis process} (c). The perception and terrain analysis can be either operated online in an onboard computer on a real-time data stream or offline using the recorded data.

\begin{figure}[h]
    \centering
    \includegraphics[width=.48\textwidth]{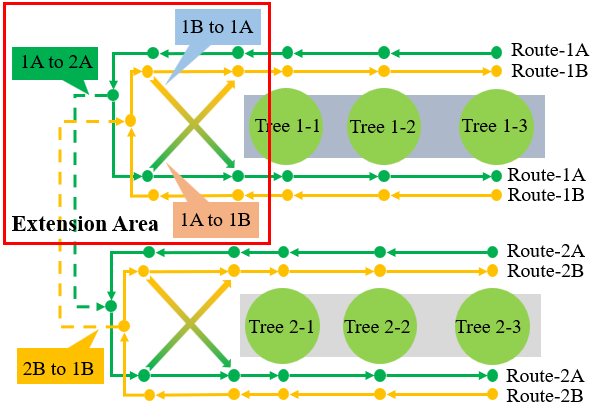}
    \caption{Accessible corridors.}
    \label{fig: navigation_basic}
\end{figure}
\textbf{Visibility graph}: Following the generation of perception and terrain analysis results, the objective of this step is to create a navigation map that provides semantic information to guide the autonomous operation of the robot within the orchard. To achieve this, we delineated terrain areas by using the Hough line detection algorithm to identify tree columns based on the position of each tree on the map, thus defining corridors between adjacent tree columns suitable for robot traversal (Figure \ref{fig: Traversability analysis process} (d)). Recognising the impracticality of performing U-turns in narrow corridors between rows of trees, we imposed a one-way travel restriction in each corridor. For each tree, two access points with different forward directions were designated on either side of the tree column (Figure \ref{fig: navigation_basic}). Nodes with the same direction were connected to form loops around the tree column. With two routes for each of the three columns and the ability for the robot to switch routes at column extension areas, as well as plan movement between different columns, the system facilitated flexible navigation. In structured orchard environments, where obstacles are more likely to be people and other vehicles, the proposed visibility graph serves as high-level guidance for path planning and obstacle avoidance during autonomous vehicle operation, allowing smooth navigation between tree rows. Using semantic information, this approach transforms unstructured orchard environments into a structured graph map with nodes (via points) and edges (connections between via points), culminating in the automatic construction of an orchard navigation graph map upon completion of the mapping process.

\subsection{Implementation details}
\begin{figure}[ht]
    \centering
    \includegraphics[width=.49\textwidth]{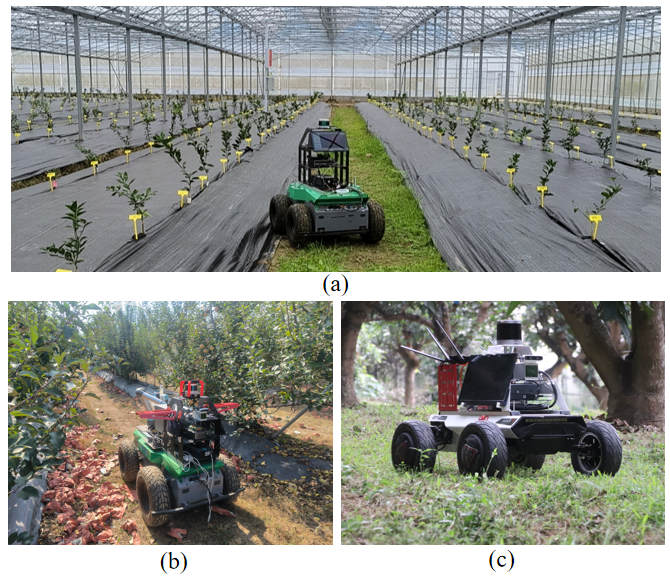}
    \caption{Mobile Robot (a) ST-100 (b) ST-HARVESTER (c) AGX-HUNTER.}
    \label{fig: hardware}
\end{figure}

\begin{figure}[ht]
    \centering
    \includegraphics[width=.49\textwidth]{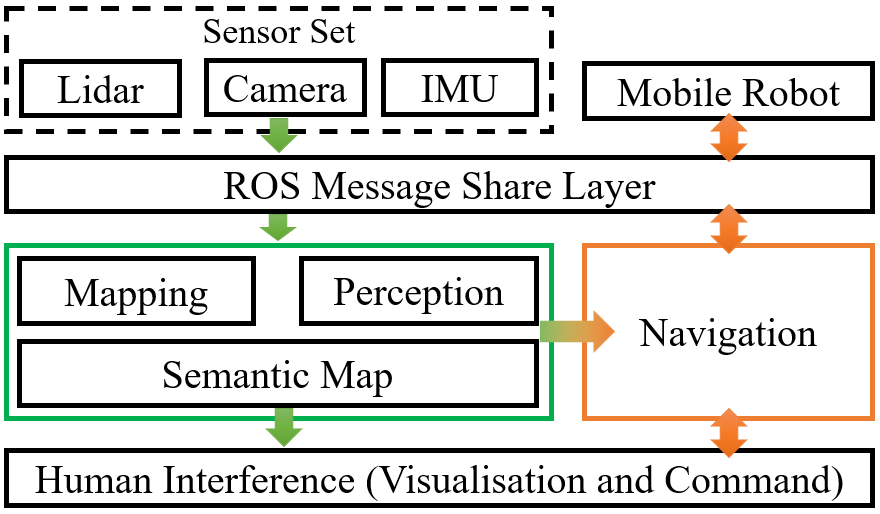}
    \caption{Software architecture of system.}
    \label{fig: software}
\end{figure}

\textbf{Hardware setup}: The overall hardware and software configuration on our developed mobile robot to perform semantic mapping of orchards are shown in Figure \ref{fig: hardware} and Figure \ref{fig: software}, respectively. Each mobile robot is equipped with a 32-line LiDAR and an external 9-axis IMU. The data acquisition frequencies of the LiDAR and the external IMU are 20 Hz and 400 Hz, respectively. 

\textbf{Software design}: The data collected by these sensors can be used to create initial point cloud maps of orchards using mapping algorithms. In our case, each robot ran the LIO-SAM algorithm. LIO-SAM utilized RS-Helios and external IMU data to construct orchard point cloud maps in XYZI format. The Realsense D435 camera is utilized for obtaining color images and is exclusively employed in this paper for visualizing color maps, without impacting the structure of the generated maps. The sensor kit on the robot is connected to the central computer (NVIDIA Xavier) using the Robot Operating System (ROS Noetic) in Ubuntu 20.04. The data transmission from the RS-Helios to the central computer is through an Ethernet port, by using rsLiDAR-ros-driver. The ROS system will hand over sensor data from the sensor set to the mapping module. The Mapping module will generate a global point cloud map, and the Perception module proposed in this article will perceive the object-level fruit tree information in the map. Through semantic mapping, a semantic rich semantic map will be generated. Finally, a visibility graph will be constructed and controlled by the ROS system for robot walking. Finally, the constructed semantic map information will be visualized with navigation information and interact with the operator to a certain extent.

\section{Experiment and Discussion} \label{section: expriment}
\subsection{Experimental setup}
\begin{figure}[h]
    \centering
    \includegraphics[width=.49\textwidth]{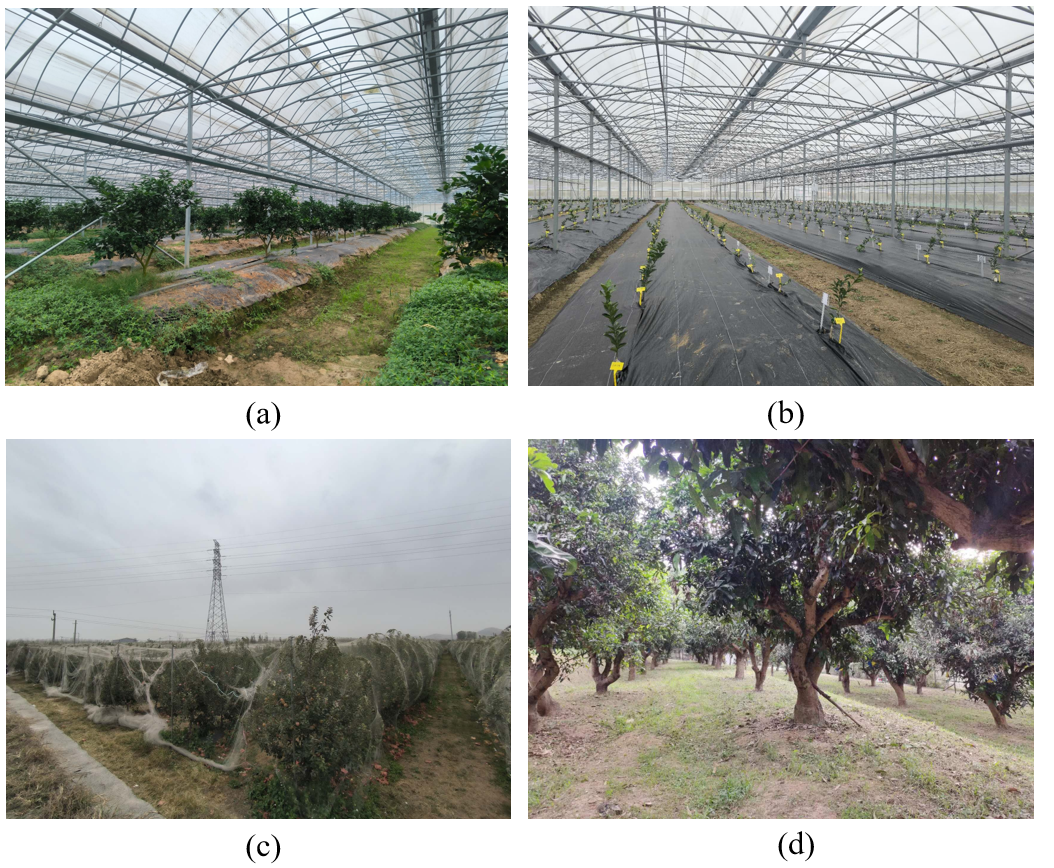}
    \caption{Experimental setting (a), (b) greenhouse  (c) unstructured apple orchard (d) outdoor lychee orchard.}
    \label{fig: Orchart+green house}
\end{figure}

\textbf{Test fields}: We conducted experiments to evaluate the proposed semantic mapping framework in a series of horticulture scenarios, including greenhouse, unstructured apple orchard, and outdoor lychee orchard (as shown in Figure \ref{fig: Orchart+green house} ).

\textbf{Experiment methods}: The first experiment analysed the effect of local maps of different sizes on the detection performance of the network. The second experiment compared the efficiency and accuracy of different feature extraction methods. Each network was trained three times and the best-performing network weight, based on verification accuracy, was saved for evaluation. We evaluated the recognition performance using mean Intersection over Union (mIoU), Mean Average Precision (MAP), and prediction time. The dataset used for training and testing the network comes from three orchard scenarios, where AGX-HUNTER, ST-100, and ST-HARVESTER mobile platforms first used the LIO-SAM algorithm to construct maps for two orchards, and obtained six point cloud maps for training the set. These maps were divided into local maps. To enhance the data, the x, y, and z directions of each local point cloud map were randomly scaled between 0 and 1, A total of 3000 enhanced local maps were obtained for use as datasets. Then, three mobile platforms mapped the third orchard and expanded the data to 500 local maps in the same way for model validation and evaluation. The third experiment analysed the effect of different parameter settings of the CSF algorithm on the universality analysis.

\subsection{Evaluation on semantic mapping}

\subsubsection{Evaluation on perception model}

\textbf{Evaluation on subdivision strategy}: Table \ref{tab:voxel-specification-strategy} shows the efficiency evaluation of the 3D-ODN model under different scales and resolutions of local maps. Resolutions of 64, 128 and 256 were tested for planarisation. Experiments compared 5$m$ x 5$m$, 10$m$ x 10$m$ and 20$m$ x 20$m$ local maps and found that larger maps resulted in longer point cloud planarisation times. The 5m x 5m map had the shortest feature extraction time (average 0.04403s), while the 20m x 20m map had the longest (average 0.6522s). The 10m x 10m map achieved the highest Intersection over Union (IoU), indicating improved detection accuracy. Smaller maps with the same planarisation resolution represented smaller areas per voxel, increasing the prominence of fruit tree features and overall detection accuracy. Notably, the 5m x 5m map showed a reduced IoU due to voids in the converted feature map, affecting the accuracy of network prediction.

\begin{table}[ht]
\centering
\caption{Voxel Specification Strategy}
\label{tab:voxel-specification-strategy}
\begin{tabular}{cccc}
\hline
strategy        & feature map time (s) & Predict time (s) & mIoU  \\
\hline
5m×5m+64×64     & 0.0398               & 0.0168            & 0.809 \\
10m×10m+64×64   & 0.1583               & 0.0168            & 0.798 \\
20m×20m+64×64   & 0.6275               & 0.0168            & 0.772 \\
5m×5m+128×128   & 0.0425               & 0.0268            & 0.725 \\
10m×10m+128×128 & 0.1681               & 0.0268            & 0.812 \\
20m×20m+128×128 & 0.6315               & 0.0268            & 0.796 \\
5m×5m+256×256   & 0.0498               & 0.0357            & 0.413 \\
10m×10m+256×256 & 0.1733               & 0.0357            & 0.798 \\
20m×20m+256×256 & 0.6975               & 0.0357            & 0.811 \\
\hline
\end{tabular}
\end{table}

Tests 2, 5, and 8 examined the impact of different resolutions on network performance, showing minimal impact on the feature extraction stage, but significant impact on the network prediction stage. Higher resolutions lead to larger feature maps, which require more parameters for prediction and affect the prediction time. Extremely high resolutions can introduce holes in fruit tree pixels, affecting prediction accuracy, while too low resolutions result in reduced feature information and recognition accuracy. To optimise both accuracy and time efficiency, a recommended configuration is a local map size of 10m x 10m with a feature map resolution of 128 x 128, striking a balance between accuracy and computational efficiency.

\textbf{Evaluation on feature encoder}: In this section, different feature extraction strategies were evaluated for their impact on the model's recognition accuracy during network training, as shown in Table \ref{tab:feature-extraction-strategy}. The View Feature Histogram (VFH) descriptor, derived from the Fast Point Feature Histogram (FPFH), was used as one method. \textbf{Test 1} The VFH was used for feature extraction by voxelising the local map along x and y directions, calculating the VFH within each voxel, resulting in a 308×128x128 feature map. \textbf{Test 2} adopted a proposed feature extraction method using average height, density and angle as features, generating a 3x128x128 feature map. For comparison, \textbf{Test 3} used PointNet++ with individual point labelling annotations and evaluated detection performance using Intersection over Union (IoU) calculations. These evaluations allowed a comprehensive comparison of the detection performance achieved by different feature extraction methods and networks.
\begin{table}[ht]
\centering
\caption{Feature extraction strategy}
\label{tab:feature-extraction-strategy}
\begin{tabular}{cccc}
\hline
strategy        & feature map time (s) & Prediction time (s) & mIoU  \\
\hline
vfh             & 2.13                  & 0.0592               & 0.824 \\
Proposed method & 0.1681                & 0.0268               & 0.812 \\
pointnet++      & \textbackslash{}      & 0.517                & 0.798 \\
pointnet++      & \textbackslash{}      & 0.943                & 0.824 \\
pointnet++      & \textbackslash{}      & 1.39                 & 0.843 \\
\hline
\end{tabular}
\end{table}

Tests 1 and 2 highlighted the superior representation of the fruit tree point cloud structure using VFH descriptors within voxels, resulting in improved detection performance by the 3D-ODN network compared to the proposed method. Although VFH achieved a higher Intersection over Union (IoU) of 0.824, the proposed method excelled in network efficiency at both the feature extraction and prediction stages. The computational requirements of VFH, with 308 parameters and channels, exceeded the efficiency of the proposed three-channel method, contributing to the superior efficiency of the latter. In tests 3, 4 and 5, PointNet++ with 1000, 2500 and 5000 sample points, using the MSG feature extraction method, achieved the highest accuracy. However, given the dimensions of the local map (10m x 10m), maintaining operational efficiency while accurately extracting fruit tree features remained a challenge. Reducing the number of sampling points improved operational efficiency but compromised feature extraction and prediction accuracy. Setting the number of sampling points to 2500 and 5000 improved accuracy at the expense of longer run times, resulting in a trade-off between accuracy and operational efficiency.

\subsubsection{Evaluation on traversability analysis}
\begin{figure}[h]
    \centering
    \includegraphics[width=.49\textwidth]{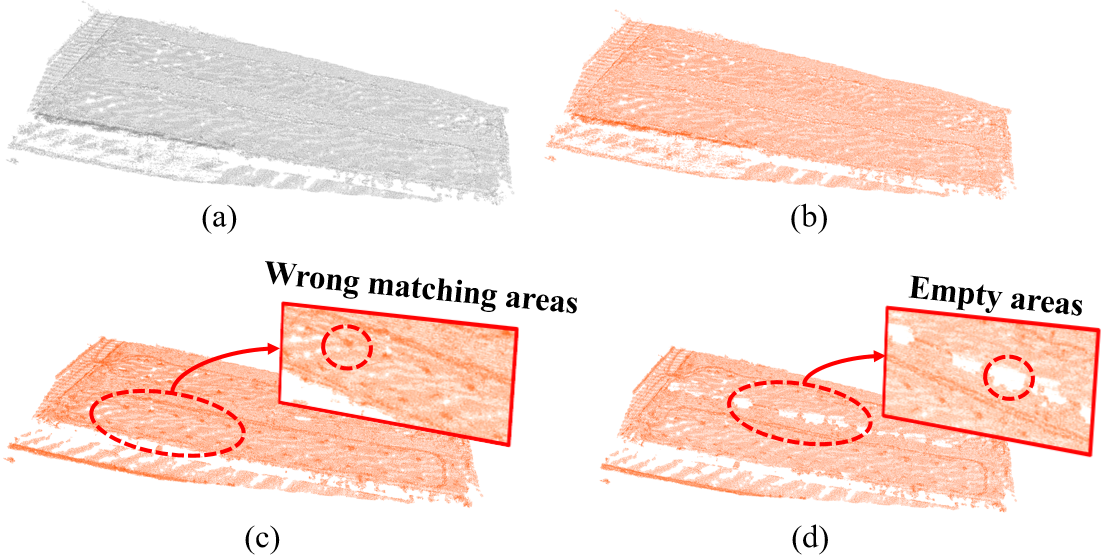}
    \caption{Comparison of CSF filtering results with different parameters. (a)ground truth (b) ideal result of CSF filtering (c) under-filtering  result by CSF (d) over-filtering result by CSF.}
    \label{fig:csf-result}
\end{figure}
Accurate ground point cloud extraction is critical for access point acquisition, especially in upland orchards with varying terrain slopes. Focusing on the orchard mapping process in upland areas, we tested the CSF algorithm as shown in Table \ref{tab: Comparison of CSF parameters}. Higher Surface Error Ratio (SER) values indicate fewer voids, while higher Surface Accuracy Ratio (SAR) values indicate more discrepancies at non-ground points. The CSF algorithm simulates a cloth surface using a particle grid to calculate the displacements caused by the force on each particle, which approximates the ground of the point cloud. It then calculates the distance between each point and the simulated cloth surface, identifying ground points below a distance threshold. To assess the impact of different parameters, we manually segmented a ground point cloud as ground truth (Figure \ref{fig:csf-result}(a)). The CSF algorithm parameters include class threshold, cloth resolution, iteration, stiffness and time step size.

\begin{table*}[ht]
\centering
\caption{Comparison of CSF algorithm parameters on ground segmentation results.}
\label{tab: Comparison of CSF parameters}
\begin{tabular}{cccccc}
\hline
Scene        & Resolution & Class Threshold ($m$) & SER & SAR & Running Time ($ms$)  \\
\hline
Lychee orchard     & 0.5       & 0.5   & 72.5\%  & 5.1\%   & 1060   \\
Lychee orchard   & 0.25         & 0.5   & 76.4\%  & 8.3\%   & 1287   \\
Lychee orchard   & 0.1         & 0.5    & 92.3\%  & 10.7\%  & 4253   \\
Lychee orchard   & 0.05        & 0.5    & 93.6\%  & 9.3\%   & 8769   \\
Lychee orchard   & 0.1         & 0.1    & 95.4\%  & 4.6\%   & 4213   \\
Lychee orchard   & 0.1         & 0.05   & 79.3\%  & 2.4\%   & 4223   \\
\hline
\end{tabular}
\smallskip
\\
\textbf{SER}: Segmentation Error Rate (SER), which indicates the proportion of misclassified points to the total number of ground trouth points.\\
\textbf{SAR}: Segmentation Accuracy Rate, which indicates the proportion of correctly classified ground points to the total number of ground trouth points.
\end{table*}

\begin{figure}[ht]
    \centering
    \includegraphics[width=.49\textwidth]{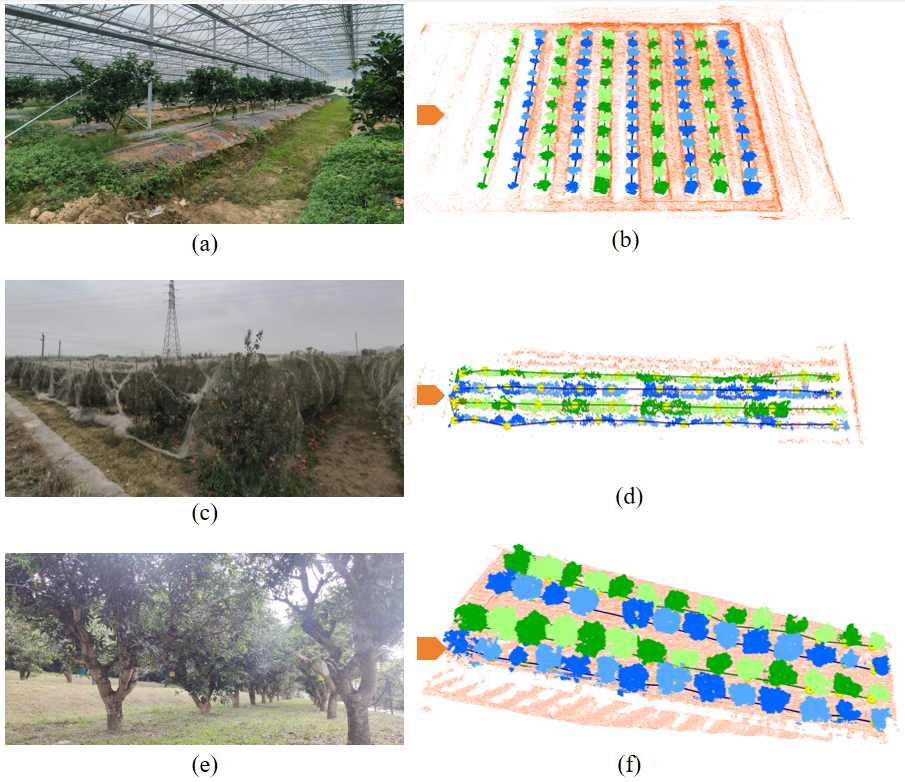}
    \caption{Demonstration of Semantic Mapping Results. (a), (c), (e) Real Scenes Captured; (b),(d),(f) Semantic results.}
    \label{fig: percep_result}
\end{figure}

In our experiments, we identified grid resolution and class threshold as the key parameters affecting segmentation results. A larger mesh resolution leads to a poor fit of simulated particles to the ground of the point cloud, resulting in voids in the extracted ground point cloud (Figure \ref{fig:csf-result}(d)). Conversely, too low a mesh resolution will result in dense simulated particles, leading to a high computational overhead (Table \ref{tab: Comparison of CSF parameters}). The class threshold, which determines the distinction between ground and non-ground points based on their distance from the simulated mesh surface, also requires careful adjustment. Too high a class threshold will result in misclassification of non-ground points (Figure \ref{fig:csf-result}(c)), while too low a threshold may result in voids in the extracted ground point cloud (Table \ref{tab: Comparison of CSF parameters}). We found optimal parameters as Resolution: 0.1, class threshold: 0.1, and kept default values for other parameters in the CSF. Combining the results of the perception module with the traversability analysis yields semantic information for the scene, as shown in Figure \ref{fig: percep_result}.

\subsection{Demonstration in applications}
\begin{table}[ht]
\centering
\caption{Comparison of the proposed visibility graph and TEB-planner global path planning.}
\label{tab: Comparison of planners}
\begin{tabular}{cccc}
\hline
Scene        &Planner & Average time (ms)   \\
\hline
Apple orchard     & Visibility graph   & 4  \\
Apple orchard   & TEB-planner & 34  \\
Greenhouse   & Visibility graph & 6  \\
Greenhouse   & TEB-planner & 45   \\
Lychee orchard & Visibility graph   & 3  \\
Lychee orchard & TEB-planner  & 31  \\
\hline
\end{tabular}
\end{table}

We tested the proposed semantic mapping and navigation framework in three scenarios, using it as an upper-level planner, with TEB-planner acting as a local planner using global path points from the visibility graph. The proposed path graph takes into account both the current direction of the vehicle and the direction on reaching each destination point, ensuring specific operational directions in orchard navigation. In contrast, the original TEB-planner does not take into account the direction of the vehicle at the end of the global path, potentially leading to opposite directions. In addition, we randomly selected 10 target points on the visibility map of each scene for comparison with TEB-planner, recording planning time (Table \ref{tab: Comparison of planners}). The visibility graph showed significantly faster computations, making it an acceptable top-level planner with minimal performance loss.

\subsubsection{Demonstration on robotic-assisted Phenotying}
\begin{figure}[h]
    \centering
    \includegraphics[width=.49\textwidth]{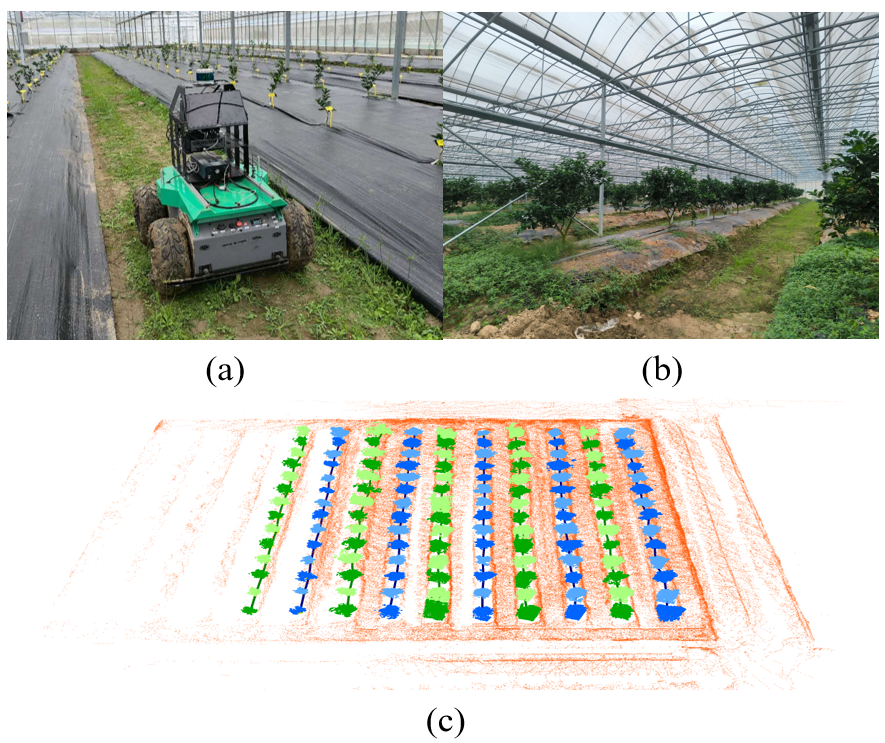}
    \caption{Demonstration on robotic Phenotying (a) ST-100 follows a global path; (b) Greenhouse Scenery (c) Semantic map.}
    \label{fig: Demonstration on robotic Phenotying}
\end{figure}
To assess the suitability of the proposed visibility graph for robotic phenotyping tasks in greenhouses, we conducted tests in a fruit tree greenhouse in Guangzhou. We compared the performance of the proposed framework with that of the TEB-planner. We used an ST-100 differential drive robot to traverse the global path derived from the visibility graph, and the resulting robotic movement within the greenhouse scene is shown in Figure \ref{fig: Demonstration on robotic Phenotying}. Due to the varying speed of the robot, we introduced slight modifications to the visibility graph. Specifically, two endpoints were strategically placed at the ends of the tree rows to allow the robot to adjust its orientation and change paths as needed. As precise orientation at the target is crucial for the phenotype acquisition task, we took the orientation of the target point into account in our tests. For a comprehensive evaluation, we compared this approach with both the native TEB-planner and a manually optimised cost map TEB-planner.

\begin{figure*}[ht]
    \centering
    \includegraphics[width=.99\textwidth]{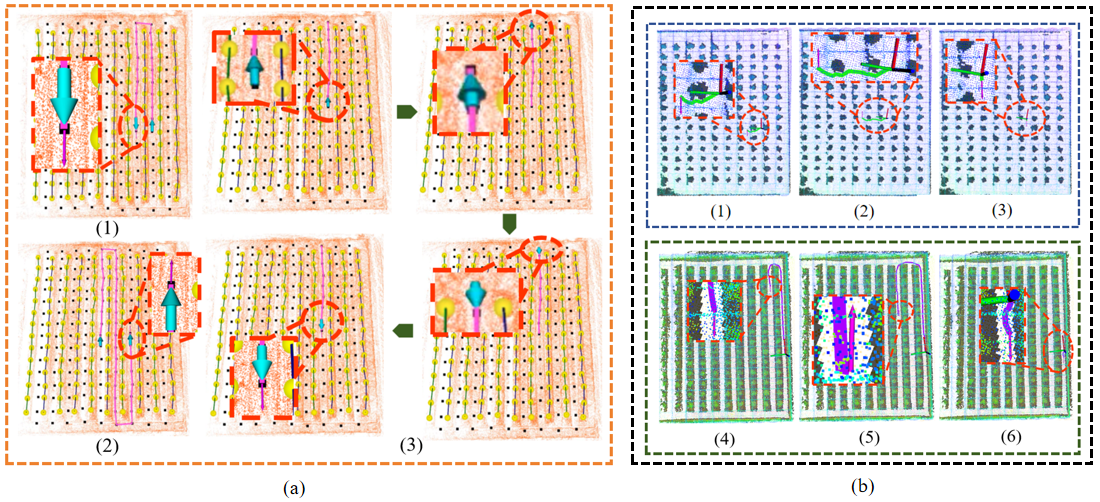}
    \caption{Visibility graph path planning results (a)(1) Case 1; (a)(2) Case 2; (a)(3) Case 3.
Raw tebplanner path planning results (b)(1) Case 1; (b)(2) Case 2; (b)(3) Case 3. TEB-planner path planning results in Manually Processed Maps (b)(4) Case 1; (b)(5) Case 2; (b)(6) Case 3.}
    \label{fig: Raw tebplanner path planning results in greenhouse}
\end{figure*}

The test was divided into three cases to evaluate the effectiveness of the planning algorithms in a greenhouse environment. In case 1, where the direction of the target point is the same as the direction of the robot, visualisations of the planning algorithms can be found in Figure \ref{fig: Raw tebplanner path planning results in greenhouse}. The TEB-planner fails to reach the target point in the specified direction and navigates through crop rows, which is considered unacceptable. Both the optimised TEB-planner and the visibility graph successfully meet the requirements. In case 2, where the direction of the target point is opposite to the robot's direction of travel, the visualisations in Figure \ref{fig: Raw tebplanner path planning results in greenhouse}(a)(2), (b)(2) and (b)(5). Neither the TEB-planner nor the optimised TEB-planner accomplishes the task, while the visibility graph adjusts the direction by travelling an additional distance. In case 3, where the target point is close to the robot but in the opposite direction, the visualisations are shown in Figure \ref{fig: Raw tebplanner path planning results in greenhouse} (a)(3), (b)(3) and (b)(6). Both TEB-planners cause the robot to rotate in place and drive to the target, an undesirable scenario in a greenhouse environment, risking vehicle collisions and crop damage. In contrast, the visibility graph guides the robot to open spaces at the ends of each aisle, allowing it to reorient itself and reach the target in required direction. From the comparison of the three cases, it can be seen that the visibility graph proposed can meet the path planning requirements of the self-driving robot in the phenotype acquisition task.

\subsubsection{Demonstration on harvesting robot}
\begin{figure}[ht]
    \centering
    \includegraphics[width=.49\textwidth]{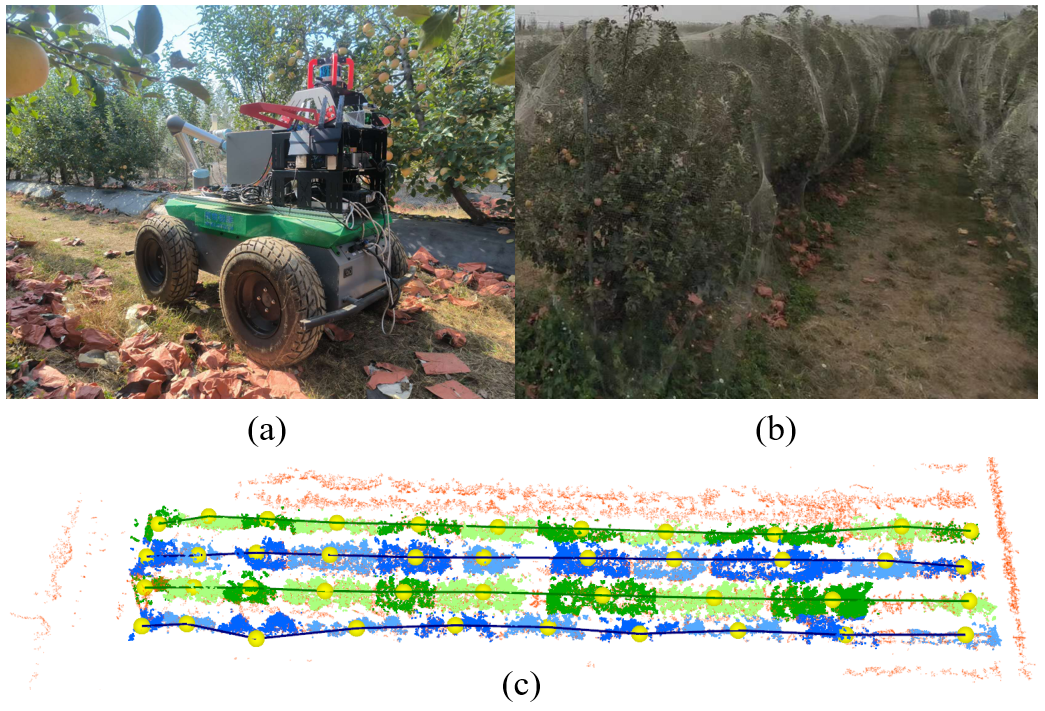}
    \caption{Demonstration on robotic harvesting (a) ST-HARVESTER (b) apple orchard Scenery (c) Semantic map.}
    \label{fig: Demonstration on robotic harvesting}
\end{figure}
In the previous section, by comparing the proposed visibility graph with TEB-planner, we learned that the visibility graph proposed in this paper is better in terms of computation time and path reasonableness, both using the original obstacle cost map and the manually constructed obstacle map. In addition, this paper also tests the visibility graph in an unstructured apple orchard scenario in Shandong Province, as shown in Figure \ref{fig: Demonstration on robotic harvesting}. 
\begin{figure}[ht]
    \centering
    \includegraphics[width=.49\textwidth]{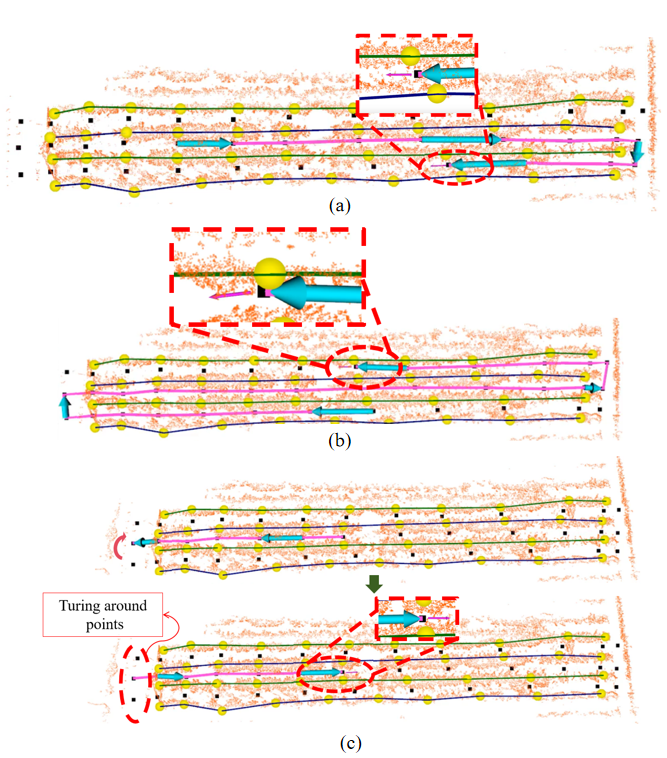}
    \caption{Visibility graph path planning results in apple orchard (a) Case 1; (b) Case 2; (c) Case 3.}
    \label{fig: visibility graph path planning test in appleorchard}
\end{figure}

It can be seen from Figure \ref{fig: Demonstration on robotic harvesting} (b) that the apple orchard has less human intervention, resulting in the growth of apple trees in a haphazard manner, which is not suitable for the perceptual framework proposed in this paper, for this reason, we artificially set up some nodes to replace the positions of fruit trees in the structured orchard. We also use the rest of the proposed framework to automatically generate access points and construct a visibility graph, due to the narrow space in the scene, we still use the visibility graph as used in the previous section, and set up turning-around points at both ends of the aisle for ST-HARVESTER to make a U-turn. and turn around points at both ends of the walkway for ST-HARVESTER to turn around and switch walkways. We test cases 1, 2, and 3 as defined in the previous section, and the path planning results are shown as (a), (b), (c) in Figure \ref{fig: visibility graph path planning test in appleorchard}. From the visualization results, it can be seen that the framework proposed in this paper can still provide reasonable and effective global path planning for self-driving vehicles after the nodes are given by humans.

\subsubsection{Demonstration of robotic monitoring in upland orchards}
Robot provide a versatile and efficient solution for surveillance and data collection in a variety of environments. Our mobile robot are equipped with SOTA sensors and mapping technology, as shown in the figure. Their efficient navigation improves the efficiency and standardization of ground data collection, thereby increasing the credibility and standardization of the data. These local data not only support precision agriculture but also contribute to digital twinning, which improves situational awareness across the board through synergistic data synergy.
\begin{figure}[h]
    \centering
    \includegraphics[width=.45\textwidth]{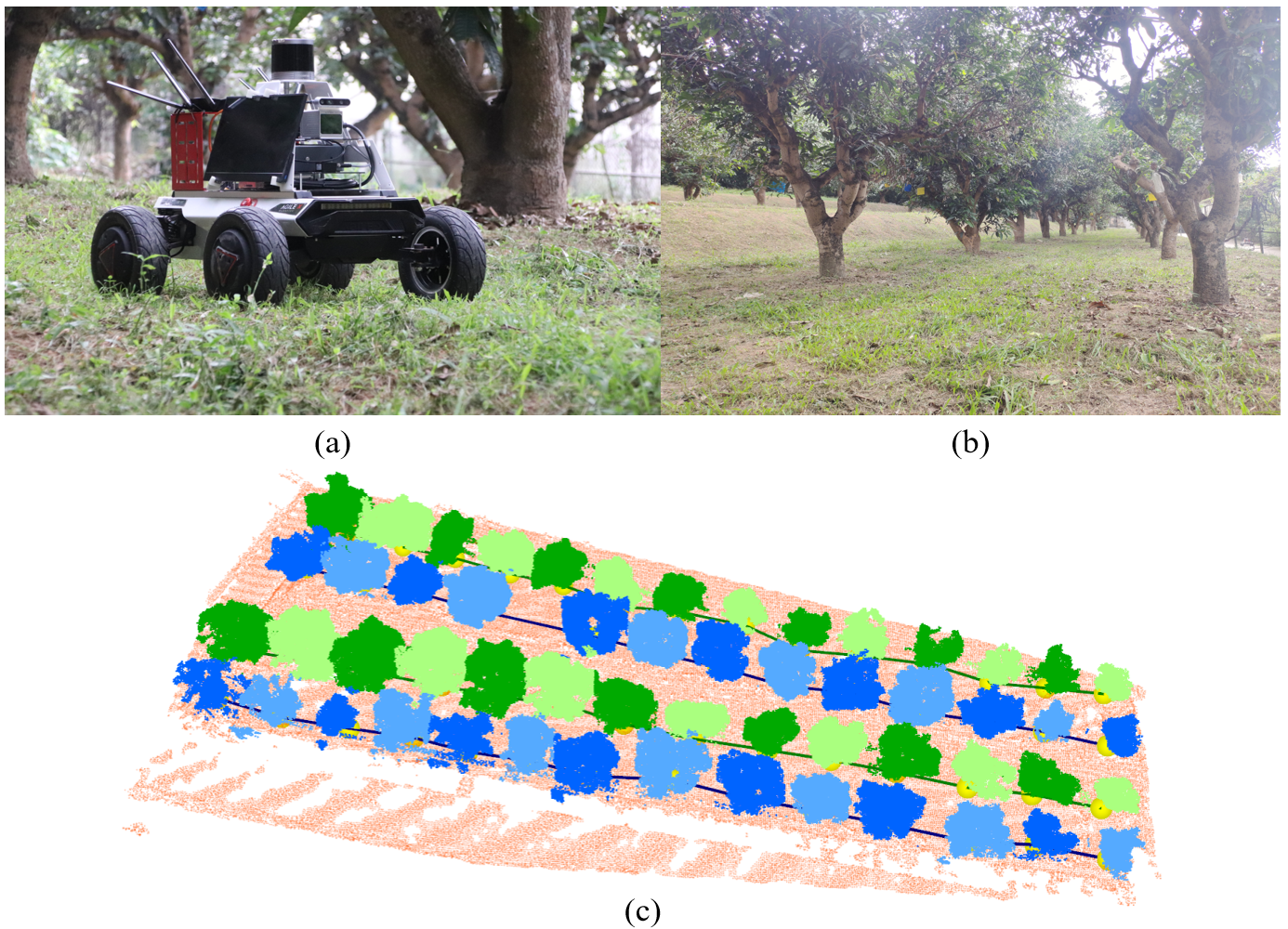}

    \caption{Demonstration on robotic monitoring (a) AGX-HUNTER follows a global path (b) Lychee orchard
Scenery (c) Semantic map.}
    \label{fig: Demonstration on robotic monitoring}
\end{figure}

\begin{figure}[ht]
    \centering
    \includegraphics[width=.48\textwidth]{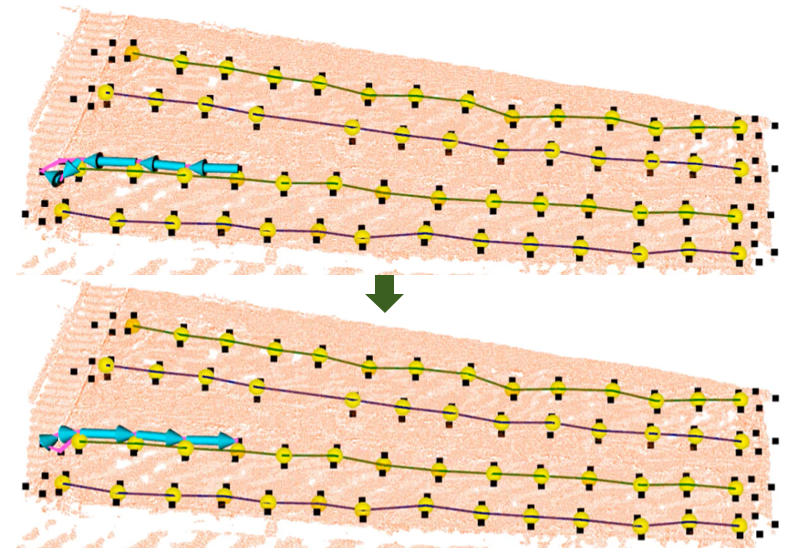}
    \caption{Demonstration on Ackerman vehicle turning around.}
    \label{fig: Demonstration on Ackermann vehicle turning around}
\end{figure}

In this study, we conducted tests on the visibility graph designed for Ackerman structure vehicles in the public lychee orchard of South China Agricultural University, as shown in Figure \ref{fig: Demonstration on robotic monitoring}. Ackerman structure vehicles face the challenge of changing their driving direction by 180 degrees during operation. Therefore, our focus was to evaluate the ability of the proposed visibility graph to adjust the direction, and the visualisation results are shown in Figure \ref{fig: Demonstration on Ackermann vehicle turning around}. The proposed visibility graph guides the vehicle to a designated area for a U-turn, significantly improving the accessibility of the path. This capability meets the monitoring requirements of Ackerman's vehicles in an orchard environment.

\subsection{Discussion}
In this study, we present a single-stage perception and semantic mapping framework for orchard environments that leverages deep learning. The framework consists of two distinct phases: perception and semantic mapping. In the perception phase, the LiDAR-generated orchard map is segmented into local map blocks, and feature extraction is performed on each block, resulting in a three-channel feature map. A key challenge is to select features that balance prediction speed and accuracy; we address this by using average height, density and the angle between the principal direction and the z-axis of the point cloud, resulting in a feature map. By experimenting with different feature extraction methods, we found that the VFH method provides favorable network predictions, but operates at an unacceptable time of 2.13 seconds. In contrast, our proposed method using density, mean height, and angle features achieves a significantly shorter processing time of 0.1949 seconds. Further experiments explored optimal settings for local map size and feature map resolution. The results show that a local map size of 10m x 10m and a feature map resolution of 128 x 128 gives the highest detection accuracy, with a mIoU of 0.812. This demonstrates the ability of our method to detect fruit trees directly from 3D point clouds while striking a balance between performance and accuracy. Our contribution lies in the efficiency of feature extraction and the optimisation of settings for superior orchard mapping.

In the semantic mapping stage, we extract the spatial distribution of fruit trees based on the perception results. Tree rows are detected using iterative Hough's linear transform, and high-dimensional semantic information, including the visibility graph, is constructed. Each step of the semantic mapping is visualised to validate the results. The comparison of the constructed visibility graph with TEB-planner in experiments demonstrates the superiority of our approach in terms of global path planning time. The global path visualisation further confirms that our visibility graph plans a more reasonable path, ensuring that self-driving vehicles follow operational requirements and driving directions, which is essential for orchard operations and phenotypic data collection.

Despite the demonstrated efficiency and accuracy of our single-stage perception and semantic mapping framework in orchard environments, there are inherent limitations. While our proposed method using density, mean height, and angle features successfully reduces the processing time to 0.1949 seconds, there is a need for further exploration to improve the efficiency of feature extraction. In addition, although optimising the local map size and feature map resolution improves detection accuracy, adaptable settings may be required due to the variation in orchard environments. Future research should prioritise the development of a more adaptive and robust framework that can perform effectively in different orchard scenarios. While our visibility graph outperforms TEB-planner, continued refinement and adaptation of the framework to different orchard conditions and operational requirements is essential to address potential limitations. Further research should extend the proposed semantic perception and mapping approach to different agricultural environments, such as vineyards and fields, to validate its applicability and generalisation. This extension will allow for comprehensive testing and refinement, taking into account the unique challenges and requirements of different agricultural environments, ultimately contributing to a more versatile solution in agricultural automation.

\section{Conclusion} \label{section: conclusion}
The proposed semantic mapping method addresses the challenges of accurate fruit tree detection and terrain categorisation in orchard environments, contributing to autonomous vehicle navigation technology. By extracting high-dimensional semantic information, including fruit tree spatial distribution, tree row details, and passable areas, the method combines efficient feature extraction and network structure to improve detection accuracy and efficiency. The method is tested in different orchard scenarios, including greenhouses, apple orchard and outdoor lychee orchards, and its effectiveness is evaluated using metrics such as mIoU, map size, and detection time. The results highlight the impact of both the map size of the point cloud map block features and the map size of the feature extraction on the accuracy and speed of network detection. A balanced choice of 10m × 10m map size and 128 × 128 feature map resolution provides an optimal balance between speed and accuracy. Experiments and visualisations in different orchard scenarios, including unstructured orchards, verify the effectiveness and generality of the method. Furthermore, a comparison with the tebplanner path planning algorithm underlines the significant advantages of the method for guiding self-driving cars in orchard environments. In conclusion, the proposed method provides an efficient and accurate solution for semantic perception and mapping in both structured and unstructured orchard point cloud maps. The constructed visibility graph provides effective global navigation guidance for self-driving vehicles in orchards, which is essential for tasks such as automated orchard harvesting, phenology collection and automated monitoring.

% \end{frontmatter}
\section*{Acknowledgement}
We thanks the funding support from CSIRO’s CERC program, Guangzhou Science and Technology Plan Project (2023B01J0046). 

\section*{Declaration of Competing Interest}
The authors declare that they have no known competing financial interests or personal relationships that could have appeared to influence the work reported in this paper.

\bibliographystyle{IEEEtran}
\bibliography{root}
\end{document}